\documentclass[runningheads]{llncs}
\usepackage[T1]{fontenc}
\usepackage{graphicx}
\usepackage{hyperref}
\usepackage{amssymb}
\usepackage{amsmath}
\usepackage{graphicx}
\usepackage{subcaption}
\usepackage{svg}
\usepackage{tikz}
\usepackage{pgfplots}
\usepackage{bm}
\usepackage{tikzscale}

\pgfplotsset{compat=1.18} 
\usepackage{color}

\newcommand{\R}{\mathbb{R}}
\newcommand{\reference}{\mathit{ref}}
\newcommand{\outlier}{\mathit{outlier}}
\newcommand{\inlier}{\mathit{inlier}}
\newcommand{\robust}{\mathit{robust}}

\DeclareMathOperator{\GS}{\textrm{GS}}
\DeclareMathOperator{\rGS}{\textrm{rGS}}
\DeclareMathOperator{\erf}{\textrm{erf}}
\DeclareMathOperator{\M}{\textrm{M}}
\DeclareMathOperator{\MSS}{\textrm{MSS}}
\DeclareMathOperator{\BSS}{\textrm{BSS}}
\DeclareMathOperator{\SSS}{\textrm{SSS}}
\DeclareMathOperator{\RSS}{\textrm{RSS}}
\DeclareMathOperator{\CSS}{\textrm{CSS}}
\DeclareMathOperator{\HIS}{\textrm{HIS}}
\DeclareMathOperator{\HM}{\textrm{H}}
\DeclareMathOperator{\T}{\textrm{T}}
\DeclareMathOperator{\D}{\textrm{D}}

\begin{document}
\title{Robust Statistical Scaling of Outlier Scores: Improving the Quality of Outlier Probabilities for Outliers (Extended Version)}
\titlerunning{Robust Statistical Scaling of Outlier Scores}
\author{Philipp R\"{o}chner\inst{1}\orcidID{0000-0003-3376-6670} 
\and Henrique O. Marques\inst{2}\orcidID{0000-0002-8273-5814} 
\and Ricardo J. G. B. Campello\inst{2}\orcidID{0000-0003-0266-3492} 
\and Arthur Zimek\inst{2}\orcidID{0000-0001-7713-4208} 
\and Franz Rothlauf\inst{1}\orcidID{0000-0003-3376-427X}
}
% \author{anonymized for double-blind review}
%
% \authorrunning{anonymous et al.}
\authorrunning{R\"{o}chner et al.}
\institute{Johannes Gutenberg University Mainz, Germany, \email{roechner@uni-mainz.de, rothlauf@uni-mainz.de} \and University of Southern Denmark, Denmark, \email{oli@sdu.dk, campello@imada.sdu.dk, zimek@imada.sdu.dk}}
\maketitle              % typeset the header of the contribution
\begin{abstract}
Outlier detection algorithms typically assign an outlier score to each observation in a dataset, indicating the degree to which an observation is an outlier. 
However, these scores are often not comparable across algorithms and can be difficult for humans to interpret.
Statistical scaling addresses this problem by transforming outlier scores into outlier probabilities without using ground-truth labels, thereby improving interpretability and comparability across algorithms. 
However, the quality of this transformation can be different for outliers and inliers.
Missing outliers in scenarios where they are of particular interest {---} such as healthcare, finance, or engineering {---} can be costly or dangerous. 
Thus, ensuring good probabilities for outliers is essential.
This paper argues that statistical scaling, as commonly used in the literature, does not produce equally good probabilities for outliers as for inliers. 
Therefore, we propose robust statistical scaling, which uses robust estimators to improve the probabilities for outliers. 
We evaluate several variants of our method against other outlier score transformations for real-world datasets and outlier detection algorithms, where it can improve the probabilities for outliers.
\end{abstract}

\let\thefootnote\relax\footnote{This is an extended version of an original article first published in \cite{rochner2024robust} by Springer Nature.}
% This is an extended version of an original article accepted for publication in SISAP 2024 by Springer Nature.

\section{Introduction}

Outlier detection algorithms compute real-valued outlier scores to identify outliers, which are observations that are significantly different from the other observations in the dataset, the so-called inliers~\cite{DBLP:books/sp/Hawkins80}. 
The rare outliers may be produced by a different mechanism than the inliers, in which case they are often referred to as anomalies~\cite{DBLP:books/sp/Hawkins80,barnett1994outliers}.
Alternatively, outliers and inliers may result from the same underlying mechanism.
In both cases, the outliers or anomalies are less likely to occur than the inliers~\cite{grubbs1969procedures}.

The real-valued outlier scores are often difficult for humans to interpret and are not comparable across algorithms. 
Therefore, several transformations have been proposed to convert outlier scores into \emph{outlier probabilities}~\cite{gao2006converting,kriegel2011interpreting,bouguessa2012modeling,bauder2017estimating,perini2021quantifying}, which quantifies the probability that an observation is an outlier~\cite{rochner2024evaluating}. 

Outlier probabilities are commonly used.
By transforming outlier scores into outlier probabilities, one can better separate outliers from inliers~\cite{gao2006converting}, normalize the outputs of multiple outlier detection algorithms for combining them into outlier ensembles~\cite{gao2006converting,kriegel2011interpreting,bauder2017estimating}, generate biased samples for constructing sequential outlier ensembles~\cite{rayana2016sequential}, compute weights for the internal evaluation of unsupervised outlier detection results~\cite{marques2020internal,DBLP:conf/sisap/MarquesZCS22}, and quantify the sample-wise confidence of outlier detectors~\cite{perini2021quantifying}.

Good outlier probabilities are concentrated around zero and one, called sharpness, separate outliers from inliers, called refinement, and reflect the frequency of outliers for observations with a similar outlier probability, called calibration~\cite{rochner2024evaluating}. 
It is usually important to discuss the quality of the probabilities for outliers and inliers separately: 
Users are often particularly interested in outliers, where it is necessary to determine good probabilities for outliers; otherwise, the probabilities of the outliers can be misleading to users and subsequent methods. 
If, for example, outlier score transformations underestimate the probabilities of outliers, important implausible information from cancer patients could be missed~\cite{rochner2023unsupervised}. 

To our knowledge, there is a lack of research on the differences between the quality of probabilities of outliers and inliers. 
For supervised classification on imbalanced datasets, the quality of probabilities for observations from different classes can be significantly different \cite{DBLP:journals/kais/WallaceD14}. 
In general, we expect the quality of probabilities to be different for outliers and inliers because outliers are rare compared to inliers and significantly different from inliers. 

We study outlier score transformations that use only the outlier scores and no external information, such as which observation is an inlier or an outlier. 
We argue and empirically show that statistical scaling~\cite{kriegel2011interpreting}, a commonly used outlier score transformation~\cite{rayana2016sequential,marques2020internal}, computes inferior probabilities for outliers than for inliers. 
Therefore, we propose robust statistical scaling, which uses robust estimators to compute outlier probabilities.
We evaluate several variants of our method against other outlier score transformations for real-world datasets and outlier detection algorithms, where it can improve the probabilities for outliers.

The following section discusses the calibration of class probabilities in supervised learning and outlier score transformations. 
In Section~\ref{sec:problem_statement}, we define the problem of this study; in Section~\ref{section:statistical_scaling}, we discuss the limitations of statistical scaling. 
Section~\ref{section:robust_statistical_scaling} proposes robust statistical scaling using robust estimators. 
In Section~\ref{section:experiments}, we describe our experiments and discuss our results in Section~\ref{section:results}, demonstrating empirically that robust statistical scaling can improve the probabilities of outliers. 
Section~\ref{sec:conclusions} concludes our study.

\section{Related Work}

For supervised classification, improving probability estimates has been intensively studied~\cite{niculescu2005predicting}. 
Commonly used methods are Platt scaling~\cite{platt1999probabilistic}, isotonic regression~\cite{zadrozny2001obtaining,zadrozny2002transforming}, and beta calibration~\cite{kull2017beyond}.
On imbalanced datasets, standard supervised calibration approaches do not necessarily compute the class probabilities of the minority class well \cite{DBLP:journals/kais/WallaceD14}. 
Therefore, it has been proposed to create multiple balanced datasets by down-sampling the majority class, training sigmoid functions on the down-sampled datasets, and averaging the calibrated probabilities for each observation.
This improves the class probabilities of the minority class without drastically worsening the class probabilities of the majority class \cite{DBLP:journals/kais/WallaceD14}.

Unlike the calibration of class probabilities in supervised classification, we explore the unsupervised transformation of outlier scores into outlier probabilities without using ground-truth labels for outliers and inliers. 
Outlier scores can be linearly scaled to the interval~$[0,1]$, which is necessary but insufficient for outlier probabilities to be good, that is, sharp, refined, and calibrated~\cite{rochner2024evaluating}. 
Therefore, outlier score transformations, such as statistical scaling, have been proposed~\cite{gao2006converting,kriegel2011interpreting,perini2021quantifying}.

\section{Problem Statement}
\label{sec:problem_statement}
We study an $n$-dimensional real-valued dataset~$\bm{X} = \{\bm{x}_i\}_{i = 1}^N$ with $N$ observations, where~$\bm{x}_i \in \R^n$, and an outlier detection algorithm $\D_{\bm{X}}: \R^n \to \R$ with outlier scores~$\bm{S} := \left \{ s_i \right \}_{i=1}^N$, where~$s_i:= \D_{\bm{X}}(\bm{x}_i)$.

We seek a transformation $\T_{\bm{S}}: \R \to [0,1]$ of outlier scores~$\bm{S}$ into outlier probabilities~\mbox{$\bm{p} = \left\{p_i \right \}_{i=1}^N$}, where
\begin{align*}
    p_i:= \T_{\bm{S}}(s_i) = \T_{\bm{S}}(\D_{\bm{X}}(\bm{x}_i)) \in [0,1]\,,
\end{align*}
so that the probabilities~$\bm{p}$ are sharp, refined, and calibrated for outliers and inliers: 
Sharp outlier probabilities are concentrated around zero and one, refined probabilities have pure ground-truth labels for observations with similar outlier probabilities, and calibrated outlier probabilities match the fraction of outliers for observations with similar outlier probabilities~\cite{rochner2024evaluating}.

The outlier score transformation~$\T_{\bm{S}}$ is unsupervised; that is, it depends only on the outlier scores~$\bm{S}$ and has no ground truth information about whether an observation is an outlier or an inlier.

\section{Background: Non-robust Statistical Scaling of Outlier Scores} \label{section:statistical_scaling}

Statistical scaling first fits a parametric distribution to the frequency distribution of outlier scores computed by a given outlier detection algorithm on a dataset~\cite{kriegel2011interpreting}. 
Therefore, the parameters of the parametric distributions are determined from the data using, for example, the method of moments~(MoM) or maximum likelihood estimation~(MLE). 
Outlier probabilities are then calculated using a modified cumulative distribution function of the approximated frequency distribution~\cite{kriegel2011interpreting}.

In the following, we refer to statistical scaling as non-robust statistical scaling~\cite{kriegel2011interpreting}.
We also limit the following discussion to non-robust Gaussian scaling, a variant of non-robust statistical scaling that uses Gaussian distributions, but similar arguments apply to other distributions. 

\begin{definition}[Non-robust Gaussian Scaling] \label{def:non-robust_gaussian_scaling}
For an outlier score distribution~$\bm{S} \subset \R$ and an outlier score~$s \in \bm{S}$, its outlier probability using non-robust Gaussian scaling~$\GS_{\bm{S}}$ is
\begin{align} \label{eq:non-robust_gaussian_scaling}
    \GS_{\bm{S}}(s) := \max \left( 0 , \erf \left( \frac{s-\mu_{\bm{S}}}{\sigma_{\bm{S}} \sqrt{2}} \right) \right)\,,
\end{align}
with the mean~$\mu_{\bm{S}}$ and the standard deviation~(SD)~$\sigma_{\bm{S}}$ of the Gaussian distribution fitted to the outlier scores~$\bm{S}$;~$\erf$ is the \emph{error function}.
\end{definition}
The error function is a scaled and translated variant of the cumulative distribution function of the Gaussian distribution: it is sigmoid-shaped, monotonically increasing, and maps the real numbers to the interval~$]\text{--}1,1[\,$. 
For~$x \geq 0$ and a normally distributed random variable~$X$ with zero mean and a SD of~$\frac{1}{\sqrt{2}}$,~$\erf(x)$ is equal to the probability that~$X$ is in the interval~$[-x,x]$~\cite{mitzenmacher2017probability}.

According to Equation~\eqref{eq:non-robust_gaussian_scaling} and because the Gaussian error function is negative for negative arguments, all observations with outlier scores less than or equal to the mean~$\mu_{\bm{S}}$ have an outlier probability of zero. 
The error function maps the outlier scores greater than the mean~$\mu_{\bm{S}}$ to the interval~$[0,1[$.

For a Gaussian distribution, the parameters derived by the MoM are the same as those derived by MLE: the mean~$\mu_{\bm{S}}$ and the SD~$\sigma_{\bm{S}}$ of the Gaussian distribution are the sample mean and sample SD of the outlier scores~$\bm{S}$. 
The mean~$\mu_{\bm{S}}$ of the Gaussian distribution is also called the center or location; the SD~$\sigma_{\bm{S}}$ is also called the scale, spread, or dispersion.

In the following, we assume that a sufficiently well-functioning outlier detection algorithm has calculated the outlier scores such that high outlier scores correspond to outliers and low outlier scores correspond to inliers. 
We also assume that we have determined the parameters of the Gaussian distribution in Definition~\ref{def:non-robust_gaussian_scaling} using an approach sensitive to long-tailed distributions, such as the MoM or, equivalently, MLE.

Since there are typically fewer outliers than inliers, the outlier score distribution of a proper outlier detection algorithm will have a long upper tail.
This asymmetry of the outlier score distribution shifts the mean to the right because the mean is sensitive to extreme values. 
As a result, Gaussian scaling maps many outlier scores to an outlier probability of zero~(see~Equation~\eqref{eq:non-robust_gaussian_scaling}), which would be correct for inliers but incorrect for outliers.
Because the SD is also sensitive to extreme values, the scale of the Gaussian distribution that approximates the outlier score distribution is large. 
As a result, the outlier probabilities of observations with outlier scores larger than the outlier score mean increase slowly~(see Equation~\eqref{eq:non-robust_gaussian_scaling}): outliers with scores larger than the score mean may have probabilities that are too low, and inliers may have outlier probabilities that are too high.

\begin{figure*}[tb!]
    \centering
    \begin{subfigure}[b]{0.45\textwidth}
        \centering
        \includegraphics[width=\linewidth]{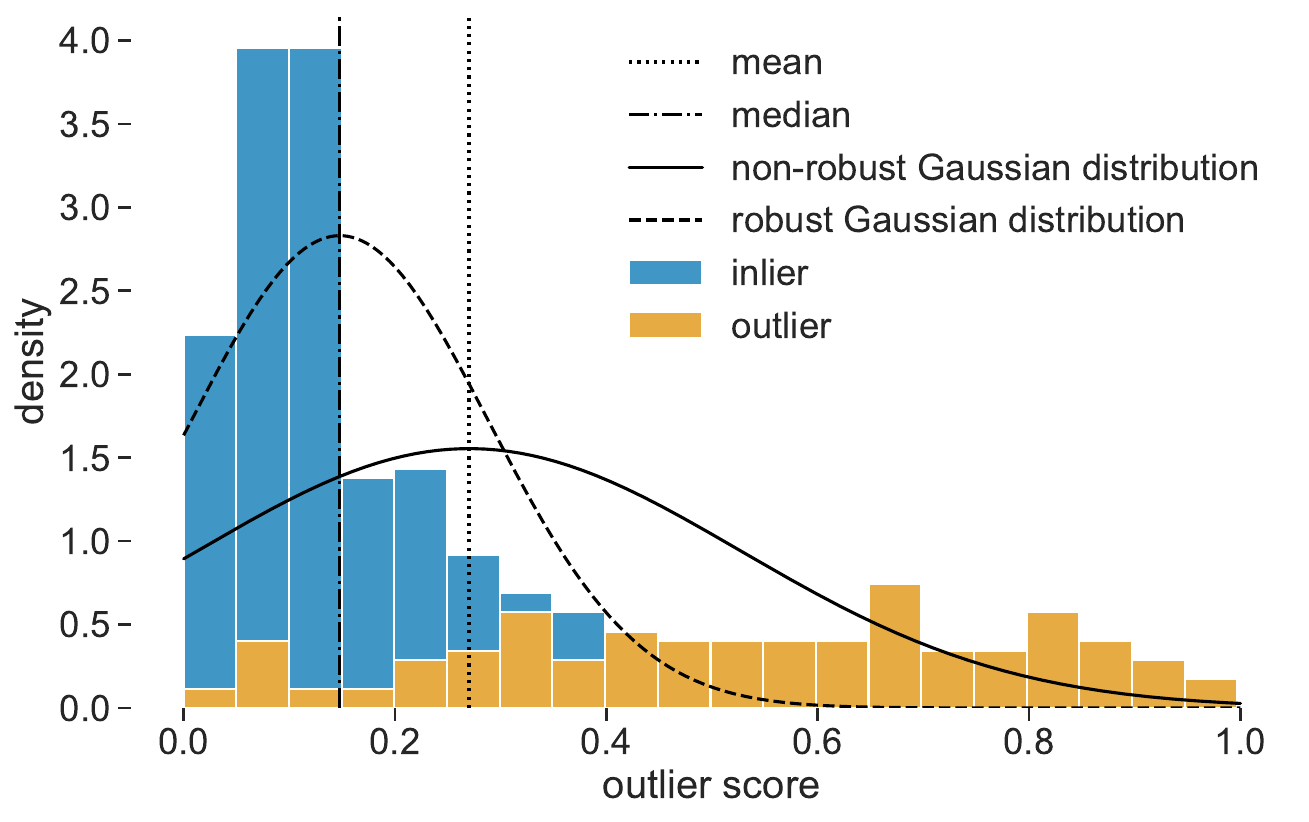}
        \caption{Outlier score empirical density distribution with non-robust and robust Gaussian density functions}
        \label{fig:outlier_scores}
    \end{subfigure}
    \hfill
    \begin{subfigure}[b]{0.45\textwidth}
        \centering
        \includegraphics[width=\linewidth]{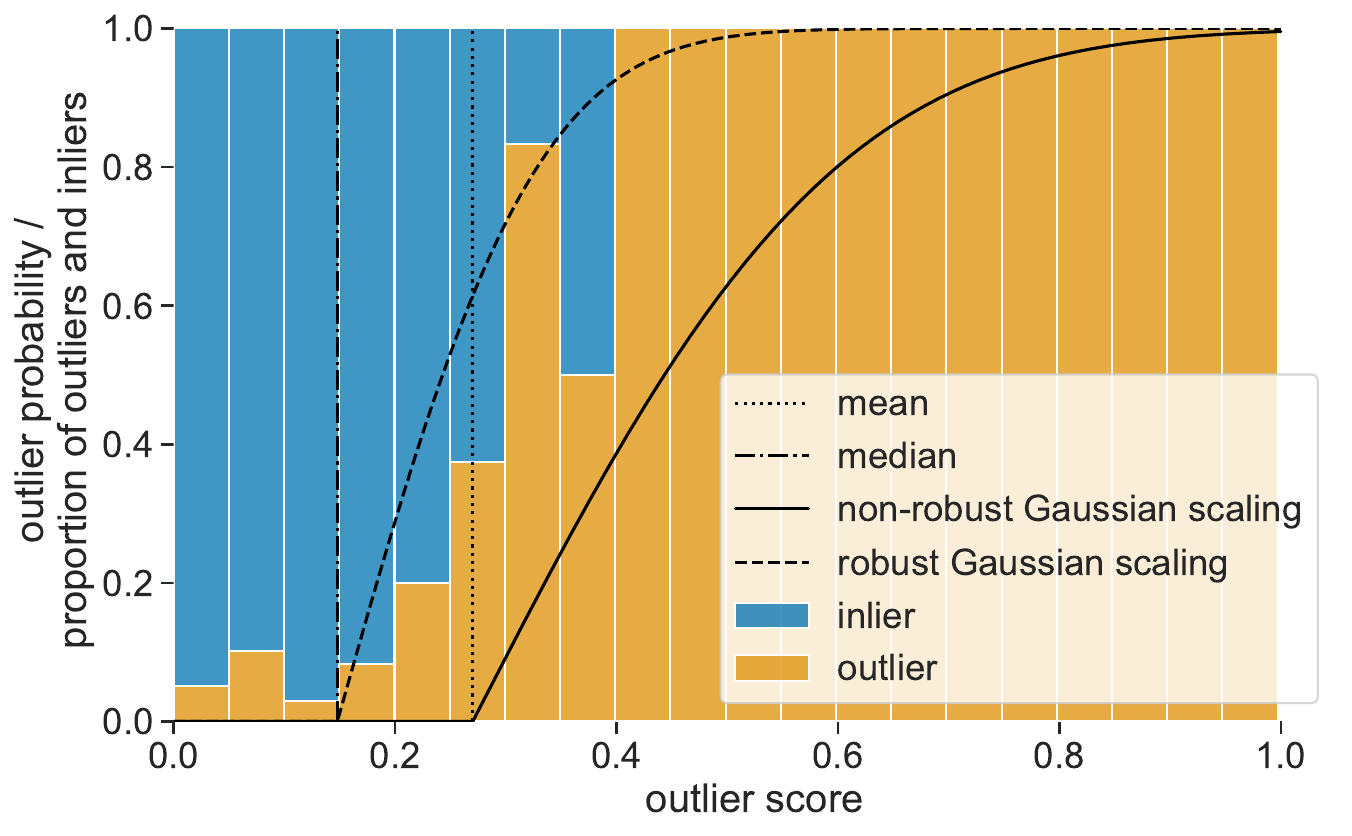}
        \caption{Transformation of outlier scores to outlier probabilities and fraction of outliers for binned outlier scores}
        \label{fig:transformation}
    \end{subfigure}
    \begin{subfigure}[b]{0.45\textwidth}
        \centering
        \includegraphics[width=\linewidth]{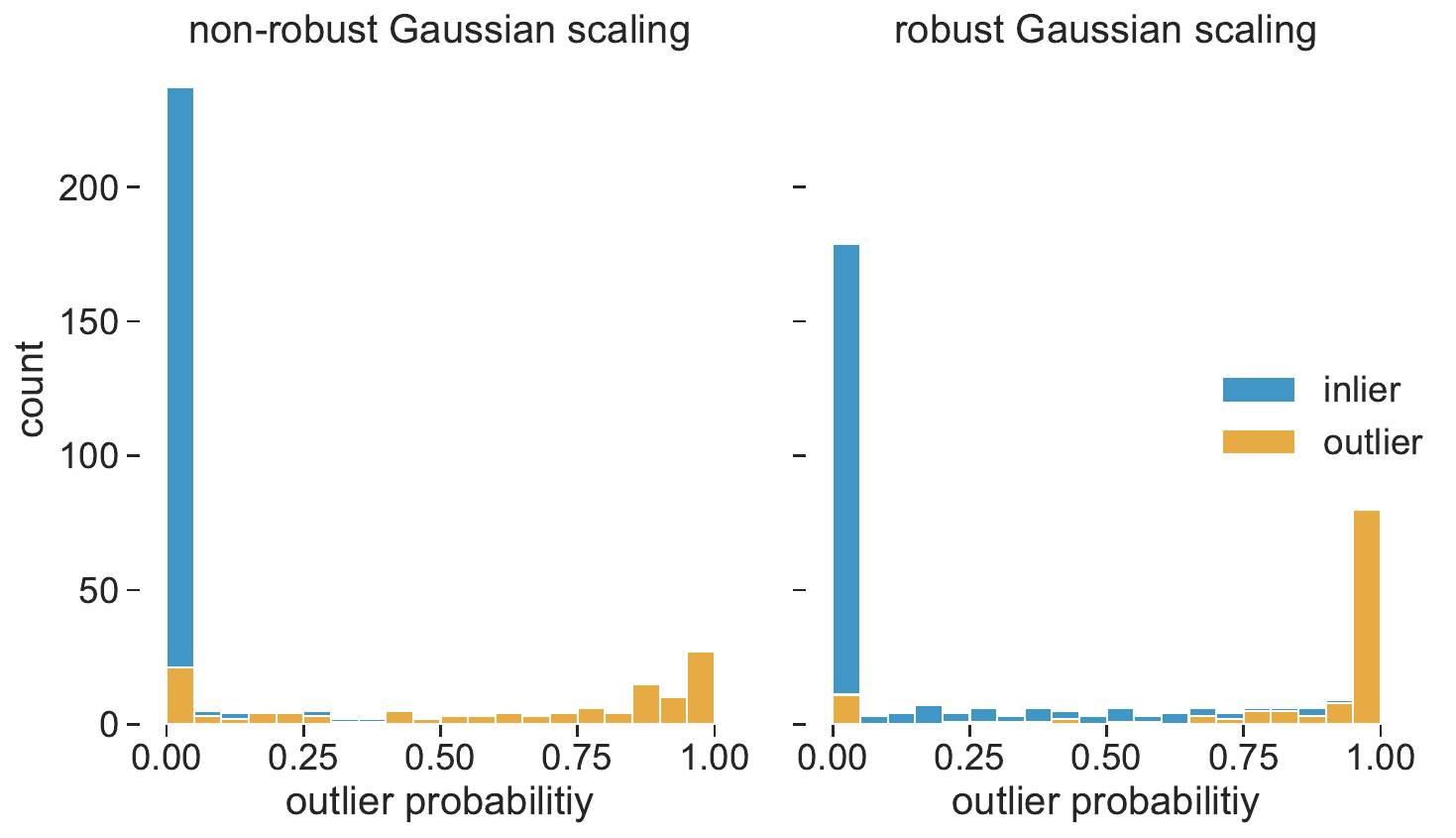}
        \caption{Empirical frequency distribution of outlier probabilities}
        \label{fig:outlier_probabilities}
    \end{subfigure}
    \hfill
    \begin{subfigure}[b]{0.45\textwidth}
        \centering
        \includegraphics[width=\linewidth]{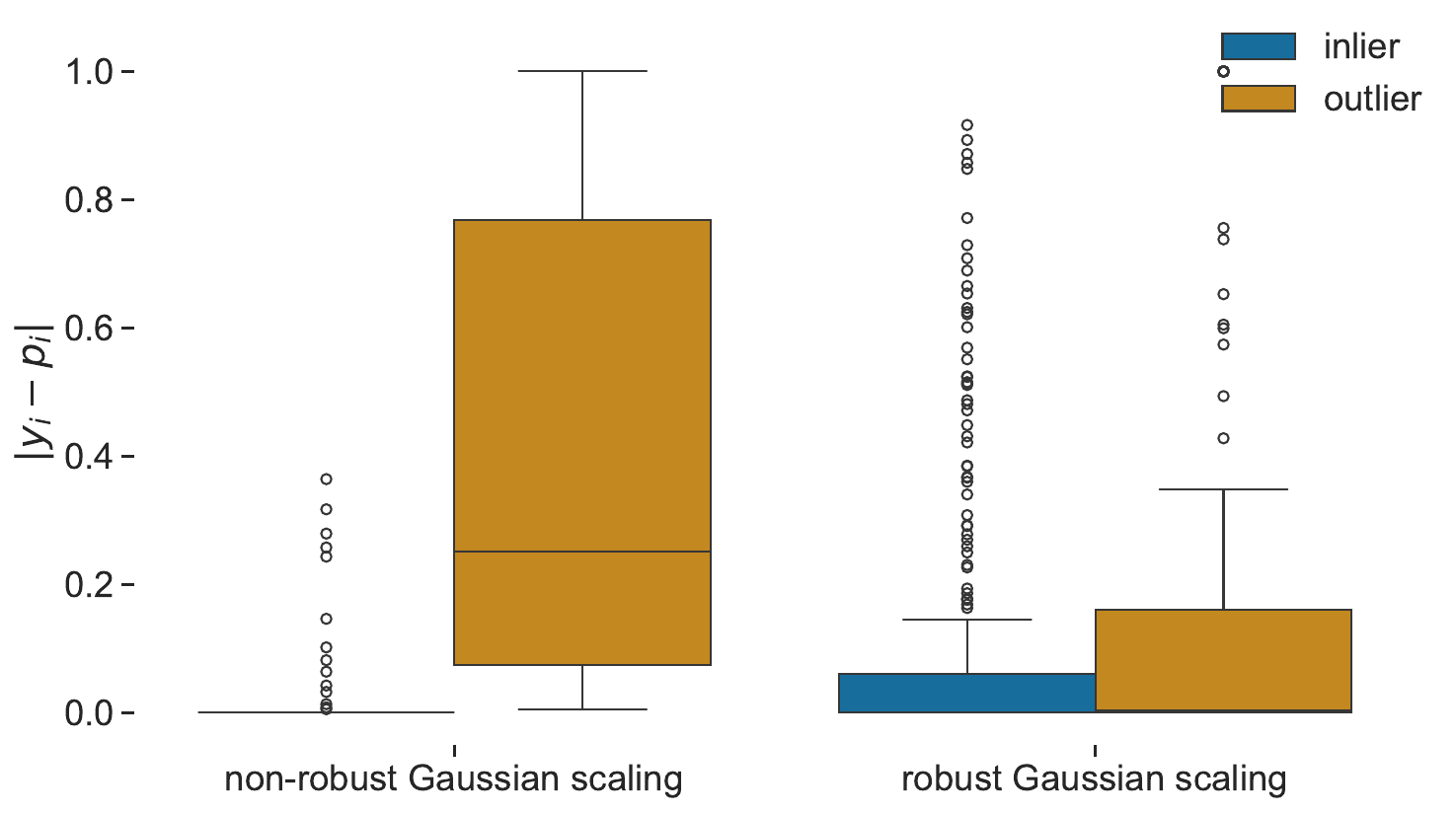}
        \caption{Residual distribution of probabilities~$p_i$ and the ground-truth labels~$y_i$
        }
        \label{fig:residuals}
    \end{subfigure}
    \caption{
    Transformation of outlier scores to outlier probabilities using non-robust and robust Gaussian scaling. 
    The $k$-Nearest Neighbors Detector~\cite{ramaswamy2000efficient} on the Ionosphere dataset~\cite{campos2016evaluation} computes the outlier scores~(Figure~\ref{fig:outlier_scores}). 
    Gaussian scaling~\cite{kriegel2011interpreting} and robust Gaussian scaling~(Figure~\ref{fig:transformation}) transform the outlier scores into outlier probabilities (Figure~\ref{fig:outlier_probabilities}). 
    We evaluate the outlier probabilities of both transformations separately for outliers and inliers~(Figure~\ref{fig:residuals}).
    In Figure~\ref{fig:outlier_scores}, the robust Gaussian density function better approximates the outlier scores of the inliers than the non-robust Gaussian density function. 
    As a result, robust Gaussian scaling reflects the proportion of outliers and inliers better, as shown in Figure~\ref{fig:transformation}. 
    In Figure~\ref{fig:outlier_probabilities}, robust Gaussian scaling correctly pushes the outlier probabilities of the outliers to one. 
    Finally, the outlier probabilities of robust Gaussian scaling have lower residuals for the outliers while slightly increasing the residuals for the inliers~(see Figure~\ref{fig:residuals}).
    }
    \label{fig:case_study}
\end{figure*}

To illustrate the effect of long-tailed outlier score distributions on non-robust Gaussian scaling~\cite{kriegel2011interpreting}, we study the $k$-Nearest Neighbors Detector~\cite{ramaswamy2000efficient} on the Ionosphere dataset~\cite{campos2016evaluation}~(Figure~\ref{fig:case_study}).
First, we examine a non-robust Gaussian distribution fitted to the outlier score distribution using the sample mean and sample SD~(solid line in Figure~\ref{fig:outlier_scores}). 
As expected, the long upper tail of the outlier score distribution shifts the non-robust Gaussian distribution to the right and increases its SD.
As a result, it does not approximate outlier scores between~$0$ and~$0.5$ well. 

Figure~\ref{fig:transformation} shows how non-robust Gaussian scaling transforms outlier scores into outlier probabilities~(solid line) according to Equation~\eqref{eq:non-robust_gaussian_scaling} and the proportion of outliers and inliers for bins with similar outlier scores. 
A calibrated transformation converts outlier scores to outlier probabilities that are close to the proportion of outliers in the corresponding bin. 
Non-robust Gaussian scaling, however, converts all outlier scores greater than the outlier score mean to outlier probabilities below the corresponding proportion of outliers.

This underestimation of outlier probabilities shifts their frequency distribution to the left.
The left part of Figure~\ref{fig:outlier_probabilities} shows the frequency distribution of outlier probabilities for non-robust Gaussian scaling. 
Many outliers have low outlier probabilities.
In contrast, almost all inliers have an outlier probability of~$0$.  

To measure the quality of the outlier probabilities of non-robust Gaussian scaling for inliers and outliers, the left part of Figure~\ref{fig:residuals} shows their residuals~\mbox{$|y_i-p_i|$} of the outlier probabilities~$p_i$ and the ground-truth labels~$y_i$, where~$y_i$ is~$0$ if the $i$-th observation is an inlier and~$1$ if it is an outlier. 
Overall, the residuals of the outliers are larger than the residuals of the inliers.

Kriegel et al.~\cite{kriegel2011interpreting} also set the mean based on a particular method's scores.
For example, LOF~\cite{DBLP:conf/sigmod/BreunigKNS00} typically computes outlier scores close to but less than~$1$ for inliers.
Setting the mean to~$1$ is then a reasonable choice. 
However, the SD remains large, and, unlike our approach, their approach is not model agnostic.
Kriegel et al.~\cite{kriegel2011interpreting} found no empirical advantages of the model-dependent approach.

\section{Robust Statistical Scaling of Outlier Scores} \label{section:robust_statistical_scaling}

We argued that non-robust statistical scaling, which fits a parametric distribution to the outlier scores in a way that is sensitive to the high scores for outliers, underestimates the probabilities of outliers. 
To improve the probabilities of outliers, we propose \emph{robust statistical scaling}, which uses robust estimators to fit a distribution to the outlier scores.
We discuss \emph{robust Gaussian scaling} as an example of robust statistical scaling, but similar arguments apply to robust statistical scaling using other distributions.
\begin{definition}[Robust Gaussian Scaling]\label{def:robust_gaussian_scaling}
    For an outlier score distribution~$\bm{S} \subset \R$ and an outlier score~\mbox{$s \in \bm{S}$}, its outlier probability using robust Gaussian scaling~$\rGS_{\bm{S}}$ is
    \begin{align} \label{eq:robust_gaussian_scaling}
        \rGS_{\bm{S}}(s) := \max \left( 0 , \erf \left( \frac{s-\mu^{\robust}_{\bm{S}}}{\sigma^{\robust}_{\bm{S}} \sqrt{2}} \right) \right)\,,
    \end{align}
    with a center~$\mu^{\robust}_{\bm{S}}$ and a scale~$\sigma^{\robust}_{\bm{S}}$ of the Gaussian distribution robustly fitted to the outlier scores~$\bm{S}$;~$\erf$ is the \emph{error function}.
\end{definition}
Because a robustly fitted Gaussian distribution is less sensitive to extreme values, its center and scale are smaller than the sample mean and sample SD. 
As a result, robust Gaussian scaling maps fewer outlier scores to outlier probabilities of~$0$.
Moreover, for outlier scores greater than the center of the outlier score distribution, the outlier probabilities increase faster. 
As a result, probabilities of outliers computed with robust Gaussian scaling are larger than those computed with non-robust Gaussian scaling.

There are several robust estimators for outlier score distributions.
For example, we can replace the sample mean~$\mu_{\bm{S}}$ and SD~$\sigma_{\bm{S}}$ in Equation~\eqref{eq:non-robust_gaussian_scaling} with robust estimates for the center and scale of the outlier score distribution.

Examples of robust estimators of a distribution's center are the median or the \emph{asymmetric trimmed mean}~\cite{maronna2019robust}. 
For the asymmetric trimmed mean, we remove a certain percentage of the largest scores from the outlier score distribution, leaving the lower tail unchanged, and calculate the mean of the remaining outlier scores. 

The \emph{normalized median absolute deviation from the median}~(nMAD), \emph{normalized interquartile range}~(nIQR), and \emph{trimmed standard deviation} are examples of robust estimators of a distribution's scale~\cite{maronna2019robust}.
The trimmed standard deviation removes large quadratic deviations between outlier scores and the distribution's center before calculating the average deviation.

In general, it is unclear how to choose how many outlier scores to trim because we have no ground-truth labels for evaluation.
We can view the amount of trimming as a hyperparameter of the outlier score transformation and could use unsupervised model selection approaches to select it~\cite{marques2020internal,zhao2022toward}.

Instead of computing a robust center and a robust scale separately, \emph{M-estimators} can robustly estimate the center and scale of distributions simultaneously~\cite{maronna2019robust}:
They iteratively estimate moments of distributions starting from initial values, such as the median for the center and the nMAD for the scale.
The center and scale estimates are then updated based on the residuals between the current center and scale estimates and the observed data, with larger residuals weighted less than smaller residuals~\cite{maronna2019robust}, which can be thought of as soft trimming~\cite{venables1997robust}.

We illustrate robust Gaussian scaling by the $k$-Nearest Neighbors Detector~\cite{ramaswamy2000efficient} on the Ionosphere dataset~\cite{campos2016evaluation}~(Figure~\ref{fig:case_study}). 
Figure~\ref{fig:outlier_scores} shows the corresponding outlier score distribution and a Gaussian distribution with the median of the outlier scores as the center and the nMAD of the outlier scores as the scale~(dashed line, which we will refer to as \emph{robust Gaussian distribution}). 
As expected, the robust Gaussian distribution is shifted to the left and approximates the inliers' outlier scores better than the non-robust Gaussian distribution. 
As a result, robust Gaussian scaling maps fewer outlier scores to an outlier probability of~$0$, and the outlier probabilities of outlier scores greater than the median increase faster compared to the non-robust Gaussian scaling~(Figure~\ref{fig:transformation}). 
Figure~\ref{fig:transformation} also shows that robust Gaussian scaling computes probabilities closer to the corresponding proportion of outliers than non-robust Gaussian scaling.
Robust Gaussian scaling correctly pushes the outlier probabilities of the outliers to one, but with the caveat that observations with outlier probabilities between approximately~$0.1$ and~$0.9$ contain more inliers~(Figure~\ref{fig:outlier_probabilities}).
Finally, the residuals of the probabilities of the outliers are lower with robust Gaussian scaling than with non-robust Gaussian scaling; the residuals of the inliers increase slightly~(Figure~\ref{fig:residuals}).

\section{Experiments} \label{section:experiments}

We first compute outlier scores for real-world datasets using outlier detection algorithms~(Section~\ref{section:data_od_algorithms}), then transform the outlier scores into outlier probabilities using outlier score transformations~(Section~\ref{section:outlier_score_transformations}), and finally evaluate their outlier probabilities~(Section~\ref{section:evaluation}).

\subsection{Datasets and Outlier Detection Algorithms} \label{section:data_od_algorithms}

We compute outlier scores on~$21$ real-world datasets \cite{campos2016evaluation}, excluding the KDD and ALOI datasets for computational reasons, using~$11$ outlier detection algorithms: Principal Component Analysis~\cite{shyu2003novel}, Kernel Principal Component Analysis~\cite{DBLP:journals/pr/Hoffmann07}, Gaussian Mixture Model~\cite{dempster1977maximum}, $k$-Nearest Neighbors Detector~\cite{ramaswamy2000efficient}, Local Outlier Factor~\cite{DBLP:conf/sigmod/BreunigKNS00}, Isolation Forest~\cite{DBLP:journals/tkdd/LiuTZ12}, Histogram-Based Outlier Detection~\cite{goldstein2012histogram}, Lightweight On-line Detector of Anomalies~\cite{DBLP:journals/ml/Pevny16}, Connectivity-Based Outlier Factor~\cite{DBLP:conf/pakdd/TangCFC02}, Outlier Detection Based on Sampling~\cite{sugiyama2013rapid}, and Unsupervised Outlier Detection Using Empirical Cumulative Distribution Functions~\cite{DBLP:journals/tkde/LiZHBIC23}.
Combining the real-world datasets and outlier detection algorithms results in~$231$ sets of outlier scores.
For all outlier detection algorithms, we used the Python library PyOD~\cite{zhao2019pyod} and set the hyperparameters to their default values. 

\subsection{Outlier Score Transformations} \label{section:outlier_score_transformations}

We investigate linear scaling~\cite{kriegel2011interpreting}, non-robust Gaussian scaling~\cite{kriegel2011interpreting}, and $10$~variants of robust Gaussian scaling~(Definition~\ref{def:robust_gaussian_scaling}) to convert the~$231$ outlier score distributions mentioned above into outlier probabilities.

For non-robust Gaussian scaling in Equation~\eqref{eq:non-robust_gaussian_scaling},~$\mu_{\bm{S}}$  is the sample mean and~$\sigma_{\bm{S}}$ the sample SD of the outlier scores~$\bm{S}$.

For robust Gaussian scaling~(Definition~\ref{def:robust_gaussian_scaling}), we determine the center of the outlier score distributions by the median and asymmetric trimmed mean, removing $10\%$ of the largest outlier scores; the scale of the distribution is computed by the nMAD and nIQR.
For both parameters, we examine the combination between all robust estimators and the non-robust sample mean and SD.
We chose~$10\%$ for trimming because it is typically larger than the percentage of outliers in a dataset and could remove the long upper tail of the outlier score distribution caused by the outliers. 
For robust Gaussian scaling with M-estimators, we use \emph{Huber's proposal 2} with \emph{Huber's T} and \emph{Tukey's biweight}, also called bisquare, as weighting functions to iteratively update the distribution's center and scale estimates depending on the residuals~\cite{huber1964robust,maronna2019robust}.

\subsection{Evaluation of Outlier Score Transformations} \label{section:evaluation}

We evaluate outlier probabilities using the Brier score, sharpness, refinement, and calibration errors and their stratified variants~\cite{rochner2024evaluating}. 
We briefly explain sharpness, refinement, and calibration in Section~\ref{sec:problem_statement}.
The Brier score measures a mixture of sharpness, refinement, and calibration.
Since inliers can dominate the above measures, the stratified measures evaluate the quality of outliers and inliers separately~\cite{rochner2024evaluating}.
All of the above measures have values between zero and one, with lower values corresponding to better outlier probabilities~\cite{rochner2024evaluating}.

In our experiments, we measure sharpness using the average entropy, refinement using the Gini index computed for bins, and calibration using the binned~$L^1$ calibration error~\cite{rochner2024evaluating}.
We compute the refinement and calibration errors using~$5$ to~$20$ equiareal bins and report the average~\cite{rochner2024evaluating}.

To compare an outlier probability distribution~$\bm{p}$ with reference outlier probabilities~$\bm{p}_{\reference}$ for an outlier probability measure~$\M$, we introduce a \emph{skill score}~\cite{murphy1973hedging}; this also makes the quality of outlier probabilities computed from different outlier score distributions better comparable.
\begin{definition}[Skill Score] \label{def:skill_score}
    For an outlier probability measure~$\M$ using ground-truth labels~$\bm{y}$, the skill score~$\MSS$ of outlier probabilities~$\bm{p}$ with respect to reference outlier probabilities~$\bm{p}_{\reference} \subset \R$ is
    \begin{align} \label{eq:skill_score}
        \MSS(\bm{p}, \bm{p}_{\reference}) 
        := \MSS(\bm{p}, \bm{p}_{\reference}, \bm{y}) 
        := -\log_2 \left( \frac{\M(\bm{p},\bm{y})}{\M(\bm{p}_{\reference},\bm{y})} \right).
    \end{align}
\end{definition}
The negative logarithm in Equation~\eqref{eq:skill_score} maps the values in the intervals~$]0,1]$ and~$[1,\infty[$ of the ratio~$\frac{\M(\bm{p},\bm{y})}{\M(\bm{p}_{\reference},\bm{y})}$ to the visually better comparable intervals~$[0,\infty[$ and~$]-\infty,0]$.

The Brier score, sharpness, refinement, and calibration errors are candidates for the outlier probability measure~$\M$ in Definition~\ref{def:skill_score}~(skill score).
For these measures, a positive skill score indicates better, a skill score of zero indicates equal, and a negative skill score indicates inferior outlier probabilities~$\bm{p}$ compared to the reference outlier probabilities~$\bm{p}_{\reference}$. 

To simultaneously evaluate the quality of an outlier probability distribution for all three measures, we aggregate the sharpness, refinement, and calibration errors into a single number. 
Although these measures theoretically have values between zero and one, they are not necessarily comparable, and we cannot aggregate them directly.
Therefore, we examine the improvement of an outlier probability distribution over a randomly selected outlier probability distribution, where we approximate the expected value of the performance of the randomly selected outlier probability distribution by the average performance of all the outlier probability distributions examined.
In other words, we first normalize the measures by dividing them by the average performance of all the outlier probability distributions examined. 
Similar to the $F_1$ score for binary classification, we then compute the \emph{harmonic improvement score} as the harmonic mean of these normalized values across different measures.
\begin{definition}[Harmonic Improvement Score] \label{def:harmonic_improvement_score}
    The harmonic improvement score of outlier probabilities~$\bm{p}$ with respect to outlier probability measures $\{\M_i\}_{i=1}^k$ and outlier probability distributions~$\bm{P}=\{\bm{p}^i\}_{i=1}^l$, with $\bm{p}^i \subset \R$ for $1 \leq i \leq l$, is 
    \begin{align*}
        \HIS(\bm{p}, \bm{P}) 
        := \HIS(\bm{p}, \bm{P}, \M, \bm{y}) 
        := \HM\left(\left\{\frac{\M_i(\bm{p}, \bm{y})}{\overline{\M_i(\bm{P},\bm{y})}}\right\}_{i=1}^k\right)\,,
    \end{align*}
    where~$\HM$ is the harmonic mean and $\overline{\M_i(\bm{P},\bm{y})}$ is the arithmetic mean of
    \begin{align*}
        \{\M_i(\bm{p}^j, \bm{y})\}_{j=1}^l.
    \end{align*}
\end{definition}
The harmonic improvement score of the sharpness, refinement, and calibration errors is positive, with smaller values indicating better outlier probabilities.

\section{Results} \label{section:results}

\subsection{Are the outlier probabilities computed by non-robust Gaussian scaling similarly good for outliers and inliers?} 

\begin{figure}[tbp!]
    \centering
    \includegraphics[width = 0.5\textwidth]{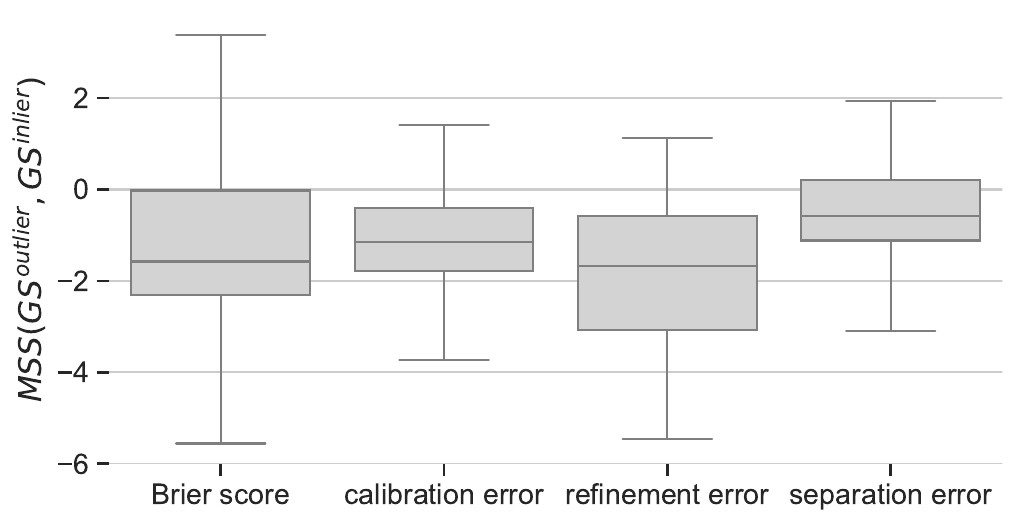}
    \caption{Skill scores~$\MSS(\GS^{\outlier},\GS^{\inlier})$ of the probabilities computed by non-robust Gaussian scaling for outliers~$\GS^{\outlier}$ compared to the probabilities for inliers~$\GS^{\inlier}$.
    A positive skill score indicates better, a skill score of zero indicates equal, and a negative skill score indicates inferior outlier probabilities~$\GS^{\outlier}$ compared to~$\GS^{\inlier}$.
    Overall, non-robust Gaussian scaling computes inferior probabilities for the outliers than for the inliers for all four measures examined.
    }
    \label{fig:figure_05_ratio_stratified_measures_non-robust_gap}
\end{figure}

First, we evaluate whether non-robust Gaussian scaling has equally good probabilities for outliers and inliers. 
Figure~\ref{fig:figure_05_ratio_stratified_measures_non-robust_gap} shows the skill score~$\MSS$ (Definition~\ref{def:skill_score}) of non-robust Gaussian scaling for the stratified Brier score, calibration, refinement, and separation errors for outliers compared to the corresponding stratified measure for inliers. 

For all four outlier probability measures examined, the distributions of the skill scores for non-robust Gaussian scaling are predominantly negative; that is, for non-robust Gaussian scaling, the stratified measures for outliers are inferior to those for inliers, consistent with our discussion in Section~\ref{section:statistical_scaling}.

\subsection{Does robust Gaussian scaling improve the probabilities of outliers compared to non-robust Gaussian scaling?}

\begin{figure*}[tbp!]
    \centering
    \begin{subfigure}[b]{0.45\textwidth}
        \centering
        \includegraphics[width = \textwidth]{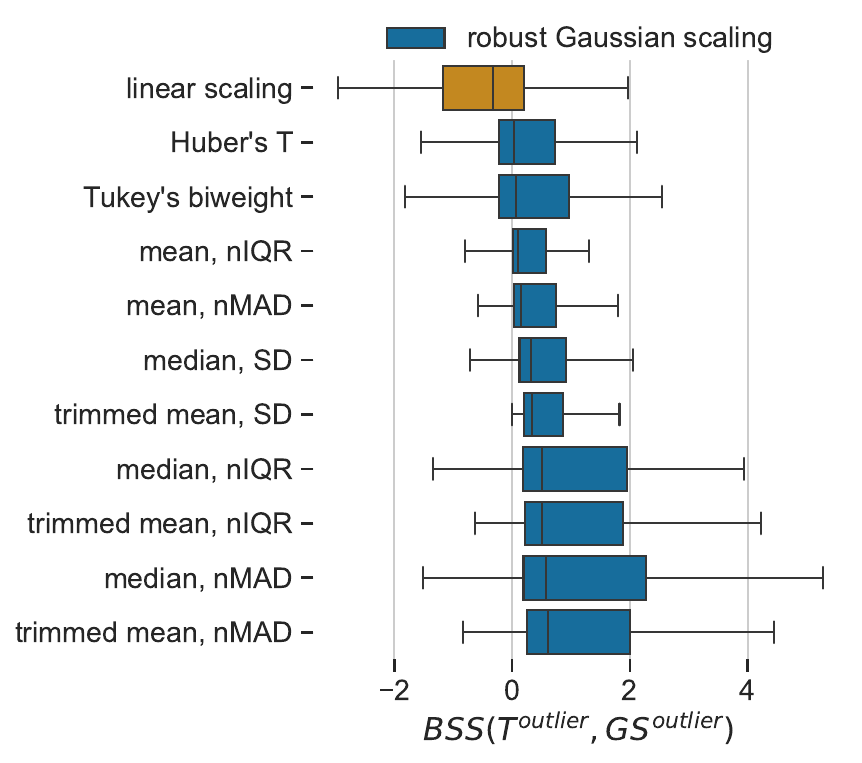}
    \caption{Brier skill score~$\BSS$}
    \label{fig:improvement_stratified_brier_score_outliers}
    \end{subfigure}
    \hfill
    \begin{subfigure}[b]{0.4\textwidth}
        \centering
        \includegraphics[width = \textwidth]{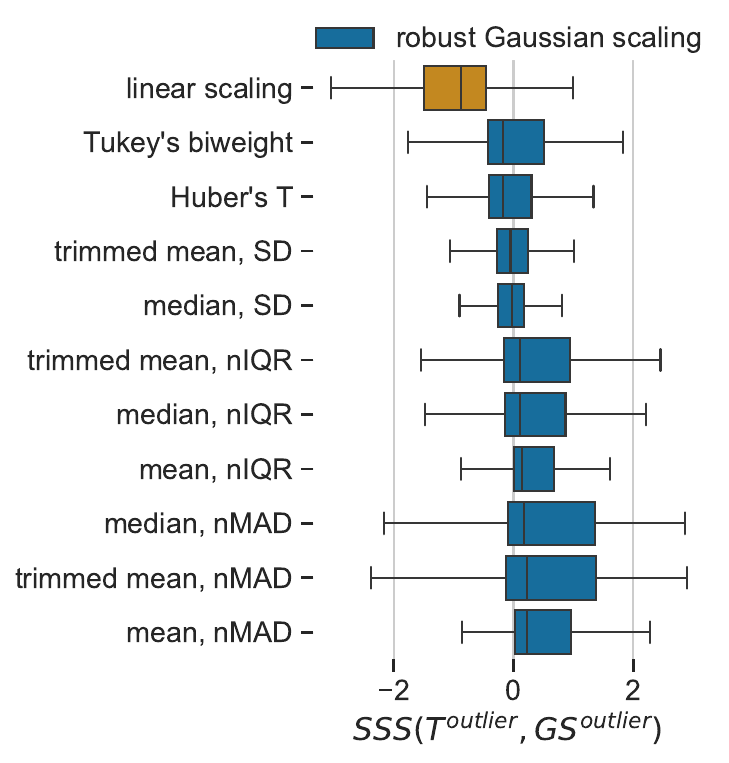}
        \caption{sharpness skill score~$\SSS$}
        \label{fig:improvement_stratified_sharpness_error_outliers}
    \end{subfigure}
    \begin{subfigure}[b]{0.45\textwidth}
        \centering
        \includegraphics[width = \textwidth]{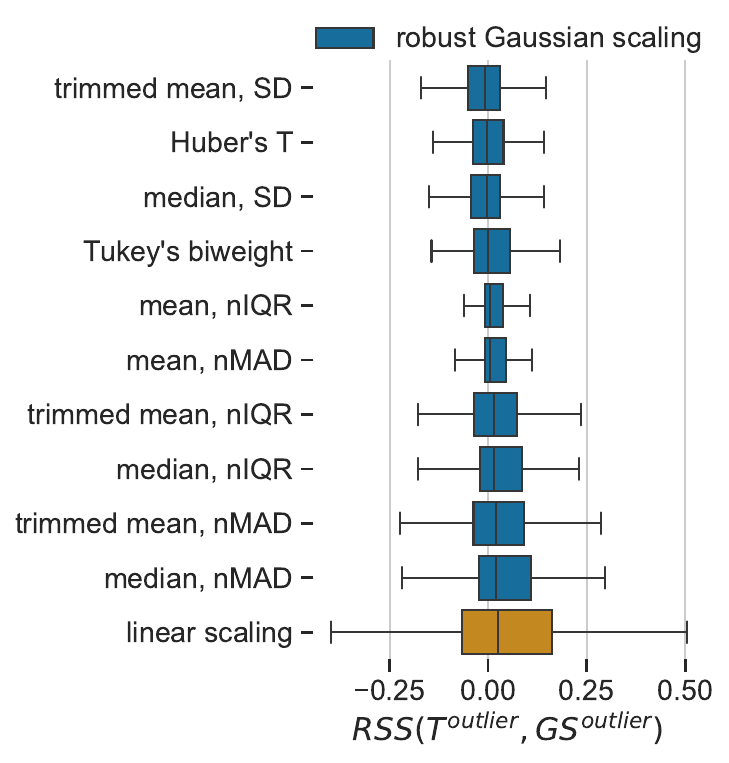}
        \caption{refinement skill score~$\RSS$}
        \label{fig:improvement_stratified_refinement_error_outliers}
    \end{subfigure}
    \hfill
    \begin{subfigure}[b]{0.45\textwidth}
        \centering
        \includegraphics[width = \textwidth]{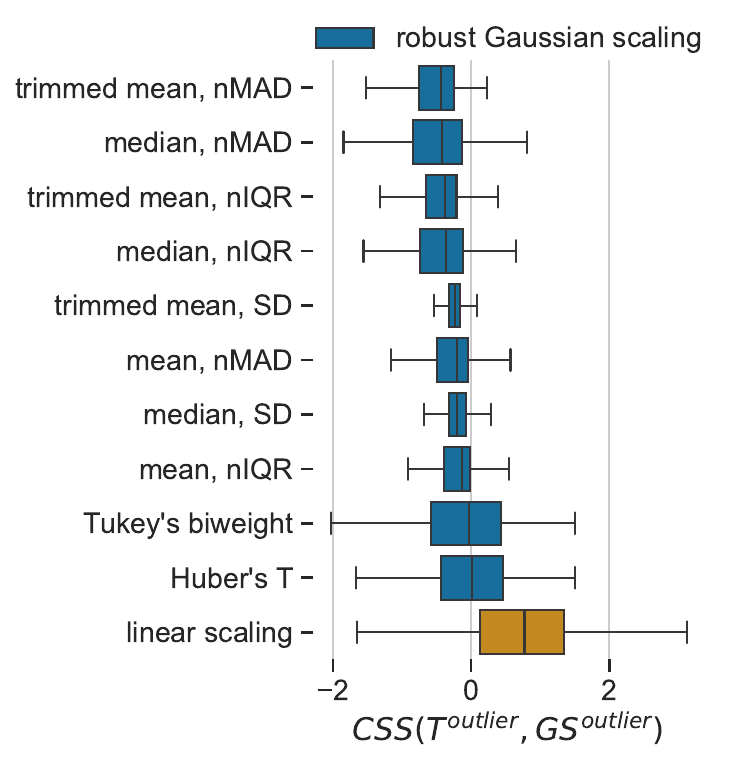}
        \caption{calibration skill score~$\CSS$}
        \label{fig:improvement_stratified_calibration_error_outliers}
    \end{subfigure}
    \caption{Skill scores~$\MSS(\T^{\outlier},\GS^{\outlier})$ of probabilities for outliers computed by outlier score transformations~$\T^{\outlier}$, which are linear scaling and variants of robust Gaussian scaling, compared to probabilities for outliers computed by non-robust Gaussian scaling~$\GS^{\outlier}$.
    A positive skill score indicates better, a skill score of zero indicates equal, and a negative skill score indicates inferior outlier probabilities~$\T^{\outlier}$ compared to~$\GS^{\outlier}$.
    Overall, all variants of robust Gaussian scaling improve the Brier score~(Figure~\ref{fig:improvement_stratified_brier_score_outliers}) for outliers compared to non-robust Gaussian scaling.
    Similarly, most variants of robust Gaussian scaling improve the sharpness~(Figure~\ref{fig:improvement_stratified_sharpness_error_outliers}) and refinement errors for outliers~(Figure~\ref{fig:improvement_stratified_refinement_error_outliers}) compared to non-robust Gaussian scaling. 
    For the calibration error for outliers~(Figure~\ref{fig:improvement_stratified_calibration_error_outliers}), the outlier probabilities of robust Gaussian scaling are inferior to the outlier probabilities of non-robust Gaussian scaling. 
    For clarity, we do not display skill scores less~(greater) than~$1.5$ times the first~(third) quartile.}
    \label{fig:gss_gaussian_outlier}
\end{figure*}

We examine whether robust Gaussian scaling improves the probabilities of outliers compared to non-robust Gaussian scaling. 
Figure~\ref{fig:gss_gaussian_outlier} shows skill scores $\MSS$ (Definition~\ref{def:skill_score}) comparing the stratified Brier score, sharpness, refinement, and calibration errors for outliers of linear scaling and variants of robust Gaussian scaling~$\T^{\outlier}$ with non-robust Gaussian scaling~$\GS^{\outlier}$. 

Compared to non-robust Gaussian scaling, all variants of robust Gaussian scaling shift the distributions of the stratified Brier skill scores for outliers above zero; this means that overall, the probabilities for outliers computed by robust Gaussian scaling have better Brier scores than those computed by non-robust Gaussian scaling.
Robust Gaussian scaling with trimmed mean as center and nMAD as scale improves the Brier score for outliers the most.

Most variants of robust Gaussian scaling improve the sharpness and refinement of probabilities for outliers compared to non-robust Gaussian scaling.
The stratified sharpness error for outliers is best for the mean as the center and nMAD as the scale for robust Gaussian scaling.

Linear scaling improves the stratified refinement and calibration errors for outliers the most compared to non-robust Gaussian scaling, followed by robust Gaussian scaling using the median and nMAD for the refinement and Huber's T for the calibration error.

As expected, robust Gaussian scaling can improve the Brier score, sharpness, and refinement of outliers compared to non-robust Gaussian scaling; only its calibration is inferior.

\subsection{Which Gaussian scaling variant best balances the overall quality of the probabilities for outliers and inliers?}

\begin{figure*}[tbp]
    \centering
    \includegraphics[width=0.6\textwidth]{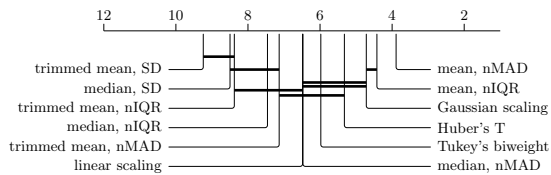}
    \caption{
    Mean rank of linear scaling, non-robust, and robust Gaussian scaling variants for the harmonic improvement score of the stratified sharpness, refinement, and calibration errors for outliers and inliers: Gaussian scaling with sample mean as center and nMAD as scale performs best.
    }
    \label{fig:cdd_hmean_sse_sre_sce}
\end{figure*}

To examine the effect of robust Gaussian scaling on the probabilities of outliers and inliers, we evaluate the probabilities when outliers and inliers, as well as sharpness, refinement, and calibration, are of equal importance.
Figure~\ref{fig:cdd_hmean_sse_sre_sce} shows the average rank of linear scaling, non-robust, and robust Gaussian scaling variants over the studied outlier score distributions for the harmonic improvement score of the stratified sharpness, refinement, and calibration errors for outliers and inliers~(Definition~\ref{def:harmonic_improvement_score}).
This measure gives equal weight to outliers and inliers, as well as to sharpness, refinement, and calibration. 
The mean ranks of the vertically connected methods are statistically indistinguishable.
As Benavoli et al.~\cite{DBLP:journals/jmlr/BenavoliCM16} have argued, we use the Wilcoxon signed-rank test~\cite{wilcoxon1945individual} instead of the Nemenyi post-hoc test~\cite{nemenyi1963distribution} to make the pairwise comparisons independent of the performance of the remaining algorithms.
We applied a Holm-Bonferroni correction to adjust the p-values~\cite{holm1979simple}.
 
Gaussian scaling with the mean as the center and nMAD as the scale estimate performs best when the probabilities for outliers and inliers, as well as sharpness, refinement, and calibration, are equally important.

\section{Conclusions}
\label{sec:conclusions}
We argue and empirically demonstrate that non-robust statistical scaling, a commonly used outlier score transformation, computes inferior probabilities for outliers than for inliers. 
Therefore, we propose robust statistical scaling using robust estimators. 
We empirically evaluate several variants of our method and show that it can improve the probabilities of outliers.
When outliers and inliers, as well as sharpness, refinement, and calibration, are equally important, statistical scaling with a non-robust center and a robust scale performed best.

This study investigated unsupervised outlier score transformations that do not use ground-truth labels.
We used, however, external knowledge about which observations are inliers and which are outliers to evaluate the outlier score transformations. 
In future work, we plan to evaluate outlier probabilities using unsupervised internal measures without using ground-truth labels for outliers and inliers.

\begin{credits}
\subsubsection{\ackname} 
The Independent Research Fund Denmark partly funded this study in the project Reliable Outlier Detection.

\subsubsection{\discintname} 
The authors have no competing interests to declare that are relevant to the content of this article.

\end{credits}
\bibliographystyle{splncs04}
\bibliography{01_robust_statistical_scaling}

\end{document}